\documentclass[11pt]{article}

\usepackage{cite}
\usepackage[utf8]{inputenc}
\usepackage[T1]{fontenc}
\usepackage{amsmath,amssymb,amsfonts}
\usepackage{algorithmic}
\usepackage{graphicx}
\usepackage{caption}
\usepackage{hyperref}
\usepackage{textcomp}
\usepackage{xcolor}
\usepackage{comment}
\usepackage[
    top=2.5cm,
    bottom=2.5cm,
    left=2.5cm,
    right=2.5cm
]{geometry}
\usepackage{booktabs}
\usepackage{tabularx}
\usepackage{authblk}
\usepackage{subcaption}

\newcommand{\keywords}[1]{%
    \par\noindent\textbf{Keywords:} #1
}

\title{Means vs. Foundation Models for Emergency Call Forecasting}

\author[1,2]{Feriel Fass}
\author[1,3]{Djemel Ziou}
\author[1,4]{Marie-Flavie Auclair-Fortier}

\affil[1]{Département d'informatique, Université de Sherbrooke,
Sherbrooke, QC, Canada, J1K 2R1}

\affil[2]{\texttt{Feriel.Fass@usherbrooke.ca} }
\affil[3]{\texttt{Djemel.Ziou@usherbrooke.ca}}
\affil[4]{\texttt{Marie-Flavie.Auclair-Fortier@USherbrooke.ca}}

\date{}
\begin{document}

\maketitle

\begin{abstract}
This study investigates whether generalized means can achieve forecasting performance comparable to that of TimeGPT for daily emergency call volume prediction. Specifically, two families of generalized means, the Hölder means (HM) and the Lehmer means (LM), are evaluated against fine-tuned TimeGPT across forecasting horizons ranging from one to seven days. Forecasting performance is assessed on a real-world emergency call dataset using multiple error metrics, and Diebold-Mariano test. Experimental results show that HM and LM achieve nearly identical predictive performance across all forecasting horizons. For short-term forecasting, both generalized means outperform TimeGPT, with reductions of mean absolute error (MAE) approximately 1.24\% and 2.48\% at horizons of one and three days, respectively. Conversely, TimeGPT achieves a lower error at the seven-day horizon, with an descresing of MAE by 1.98\%. However, Diebold–Mariano tests reveal no statistically significant differences between the methods at the evaluated horizons. These findings suggest that generalized means can provide competitive forecasting performance compared with a energyvore foundation model.
\end{abstract}

\section{Introduction}

Prehospital emergency services (PES) play an essential role in healthcare systems by providing timely medical assistance. However, increasing emergency call volumes have placed additional pressure on operational capacity and response times \cite{ccsll,fass2025survey}. Accurate call volume forecasting is therefore essential for anticipating demand, planning ambulance operations, and supporting personnel and resource allocation \cite{fass2025state,fass2025survey,fildes2002telecommunications,ibrahim2016modeling}.

Numerous forecasting approaches have been investigated, ranging from frugal models to energyvore models \cite{fass2025survey,fass2025state,fass2026TimeGPT}. Depending on operational objectives, forecasts can be produced at different temporal resolutions, including annual, day, and hour levels, and may incorporate spatial information. Several studies have also investigated explanatory variables, such as meteorological, temporal, demographic, and socioeconomic factors, to improve forecasting accuracy \cite{fass2025state}. At a daily granularity, various forecasting methods have been evaluated, including autoregressive models, count regression, ensemble learning, and neural networks. In \cite{fass2025state}, SARIMAX, Poisson regression (PR), multilayer perceptron (MLP), long short-term memory (LSTM), XGBoost, random forest (RF), gradient boosting regression trees (GBRT), and LightGBM (LGBM) were compared using temporal, meteorological, and calendar-related variables. SARIMAX achieved the lowest mean absolute percentage error (MAPE) of 6.05\%, compared with 38.14\% for XGBoost. More recently, transfer learning and foundation models have emerged as alternative forecasting methods \cite{benamirouche2026,garza2024timegpt1}. Foundation models are pretrained on large collections of time series from different domains to learn transferable temporal representations, enabling forecasting through zero-shot inference or task-specific fine-tuning. Among these models, TimeGPT has been evaluated for emergency call volume forecasting in univariate and multivariate settings \cite{fass2026TimeGPT}. However, SARIMAX outperformed TimeGPT across the evaluated scenarios, raising questions about the necessity of complex foundation models for this forecasting task.

While recent research has increasingly focused energyvore models, frugal methods may offer competitive forecasting performance. Generalized means aggregate observations using a parameter that controls thedata selection. Two such families, the Hölder means (HM) and Lehmer means (LM), generalize several classical measures of central tendency \cite{ziou2024using}. Their estimation is straighforward, no learning is required, frugal in terms of computational resources. This study investigates whether the Hölder and Lehmer generalized means (HM and LM, respectively) can achieve forecasting performance comparable to TimeGPT in predicting daily emergency call volumes over forecasting horizons ranging from one to seven days. Performance is assessed using several error metrics and Diebold–Mariano tests to evaluate the statistical significance of predictive differences.

The remainder of this paper is organized as follows. Section \ref{sec2} describes the dataset. Section \ref{sec3} presents the forecasting methods. Section \ref{sec4} reports the experimental setup, evaluation metrics, and results. Finally, Section \ref{sec5} presents the conclusions and future research directions.
\section{Data}
\label{sec2}
Experiments are conducted using real emergency call data provided by the health communication center of the Laurentides and Lanaudière regions in Quebec (CCSLL). The dataset contains approximately 600{,}000 calls collected between 2012 and 2017. Each record contains temporal attributes (e.g., call timestamps and dispatch and assignment times), spatial attributes (e.g., geographic coordinates), and contextual attributes (e.g., priority levels and clinical codes).

For daily granularity, call records are aggregated by counting the number of calls within consecutive, non-overlapping 24-hour intervals. The resulting daily time series contains 2,101 observations, with daily call volumes ranging from 179 to 469 calls, an average of $\mu=273$ calls/day, and a standard deviation $\sigma=37.36$. The dispersion factor $\sigma^2/\mu= 5.10$, indicates substantial overdispersion, suggesting that the call volumes deviate considerably from the equidispersion assumption of the Poisson distribution.

It should be  noted an important distinction arises between holidays and non-holidays. The dataset comprises 76 holiday observations and 2,025 non-holiday observations. Holidays include New Year's Day, the day after New Year's Day, Good Friday, Easter Monday, National Patriots' Day, Quebec's National Holiday, Canada Day, Labour Day, Thanksgiving, Christmas Eve, Christmas Day, Boxing Day, and New Year's Eve. During holidays, the average daily call volume increases to 291 calls/day, compared with 272 calls/day on non-holidays, corresponding to an increase of approximately 6.93\%. Moreover, holiday call volumes exhibit substantially greater variability, with a standard deviation of 52.81, compared with 36.49, respectively, for non-holidays. The corresponding dispersion factor increases from 4.88 on non-holidays to 9.55 on holidays, indicating considerably stronger overdispersion during holiday periods. This distinction aims improve forecasting performance, as suggested in \cite{fass2025state}. The non-holiday dataset was then divided chronologically into training, and test sets. Finally, note that observations from 2012 to 2016 were used for training, and observations from 2017 were reserved for testing.

\section{Methods}
\label{sec3}
Let $Y=\{Y_t\}_{t>0}$ be a discrete-time stochastic process representing emergency call demand, where $Y_t$ denotes the number of calls recorded during the interval $[t-g,t]$, where the unity means a day as the temporal aggregation granularity $g$. In addition, consider a multivariate process $X=\{X_t\}_{t>0}$, where $X_t=(X_t^1,\ldots,X_t^d)\in\mathbb{R}^d$ contains $d$ explanatory temporal, meteorological, demographic, and socioeconomic variables. Using the observations $Y_{1:t}=(Y_1,\ldots,Y_t)$ and $X_{1:t+h}=(X_1,\ldots,X_t,,\ldots,X_{t+h})$, the forecasting consists of estimating the emergency call volume at a future time $t+h$, where $h$ denotes the forecasting horizon. This can be expressed as
\begin{equation}
\label{eq:F}
\hat{Y}_{t+h} = F\left(Y_{1:t},X_{1:t+h};\theta\right),
\end{equation}
where $F$ represents the forecasting model, $\theta$ denotes its parameters.

\begin{figure*}[t]
    \centering
    \includegraphics[width=\linewidth]{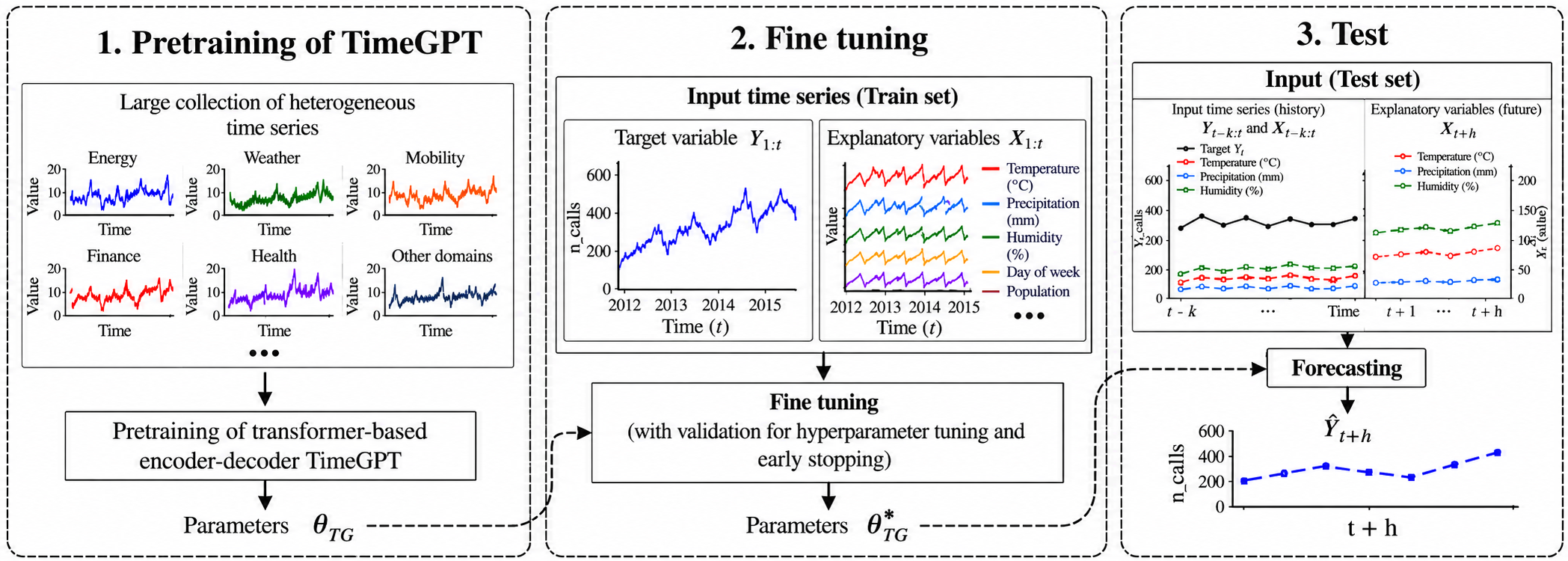}
    \caption{TimeGPT framework for emergency call forecasting, including pretraining, fine-tuning, and testing stages \cite{fass2026TimeGPT}.}
    \label{fig:timegpt}
\end{figure*}

Several methods are considered for specifying \(F\), including TimeGPT, SARIMAX, PR, NBR, RF, GBRT, XGBoost, LGBM, MLP, and LSTM. These methods, except TimeGPT, are described in \cite{fass2025state,fass2025maroc}. TimeGPT is a pretrained foundation model designed for time-series forecasting, introduced in \cite{garza2024timegpt1} and used in \cite{fass2026TimeGPT} for forecasting emergency call volume. Its architecture is based on an encoder--decoder structure adapted to sequential numerical data. Figure~\ref{fig:timegpt} illustrates the three main stages involved in the use of TimeGPT for emergency call forecasting: pretraining, fine-tuning, and testing. TimeGPT is initially pretrained on a large and heterogeneous collection of time series from multiple domains, such as energy, weather, mobility, finance, and healthcare, comprising approximately 100 billion observations. During pretraining, the model $theta_{TG}$ is estimated encoding temporal patterns shared across different time series, such as trends, seasonalities, and short- and long-range dependencies. It is the preliminary model subsequently transferred to the emergency call forecasting task. It is fine-tuned on the emergency call training set to adapt to the characteristics of the target forecasting task giving as output \(\theta_{\mathrm{TG}}^{*}\). The training input consists of the historical emergency call volumes \(Y_{1:t}\) and the corresponding historical explanatory variables \(X_{1:t}\). The parameters \(\theta_{\mathrm{TG}}^{*}\) are used for forecasting $Y_{t+h}$. The used data are the history of emergency call volumes \(Y_{t-k:t}\), the corresponding historical explanatory vectors \(X_{t-k:t}\), and the explanatory vectors available for the future forecasting time \(X_{t+h}\). To summarize, the fine-tuning process can be expressed as
\begin{equation}
\theta_{\mathrm{TG}}
\xrightarrow[\left(Y_{1:t},X_{1:t}\right)]
{\text{fine-tuning}}
\theta_{\mathrm{TG}}^{*}.
\end{equation}
For forecasting, Eq.~\ref{eq:F} is reused as follows:
\begin{equation}
\label{eq:F_TG}
\hat{Y}_{t+h}
=
F_{\mathrm{TG}}
\left(
Y_{t-k:t},
X_{t-k:t},
X_{t+h};
\theta_{\mathrm{TG}}^{*}
\right).
\end{equation}

Univariate modeling can be considered a particular case of the multivariate modeling in which forecasting relies exclusively on the historical values of the target variable, without incorporating explanatory variables. In this case, Eq.~\ref{eq:F_TG} reduces to: 
\begin{equation}
\hat{Y}_{t+h}
=
F_{\mathrm{TG}}
\left(
Y_{t-k:t};
\theta_{\mathrm{TG}}^{*}
\right),
\end{equation}
with 
\begin{equation}
\theta_{\mathrm{TG}}
\xrightarrow[\text{training data }Y_{1:t}]
{\text{fine-tuning}}
\theta_{\mathrm{TG}}^{*}.
\end{equation}

In the following, we introduce the Hölder means (HM), and the Lehmer means (LM) as additional forecasting methods \cite{ziou2024using}. Unlike parametric statistical and machine learning based approaches, both methods provide forecasting mechanisms based on the aggregation of historical observations. Similar to the univariate formulation of TimeGPT, HM and LM rely exclusively on historical emergency call volumes $Y_{1:t}$, without incorporating explanatory variables. They belong to families of central tendencies that generalize several conventional averaging methods, with each family characterized by a real-valued parameter $\alpha$ controlling the selection of observations. It has been shown that the mean is a data-selection mechanism \cite{ziou2024using}.

Let $w$ denote the historical window length and $\alpha\in\mathbb{R}$ the selection order. Both HM and LM assume strictly positive observations ($Y_t>0$). If zero-valued observations are present, a constant $c>0$ can be added to all observations, yielding the transformed series $Y_t'=Y_t+c$, to ensure positivity. Forecasts obtained on the transformed scale are then adjusted by subtracting $c$. In the present study, all observed emergency call volumes are strictly positive; therefore, no transformation is required. Given the observations available at time $t$, the predicted emergency call volume at horizon $h$ using HM is defined as

\begin{equation}
\label{eq:generalized_mean}
\hat{Y}_{t+h}=\mathrm{HM}(\alpha,w)=
\begin{cases}
\displaystyle
\exp\left(
\frac{1}{w}\sum_{i=0}^{w-1}\log Y_{t-i}
\right),
& \text{if } \alpha=0, \\[10pt]
\displaystyle
\left(
\frac{1}{w}\sum_{i=0}^{w-1}Y_{t-i}^{\alpha}
\right)^{1/\alpha},
& \text{elsewhere}.
\end{cases}
\end{equation}

The Lehmer mean provides an alternative aggregation mechanism, defined as
\begin{equation}
\label{eq:lehmer_mean}
\hat{Y}_{t+h}=\mathrm{LM}(\alpha,w)
=
\frac{
\sum_{i=0}^{w-1}Y_{t-i}^{\alpha}
}{
\sum_{i=0}^{w-1}Y_{t-i}^{\alpha-1}
}.
\end{equation}

Both families generalize several conventional central tendencies. In particular, the arithmetic mean is obtained for $\alpha=1$ in both HM and LM, while the harmonic mean corresponds to $\alpha=-1$ for HM and $\alpha=0$ for LM, and finally for geometric mean $\alpha=0$ in HM. Furthermore, for $\alpha=2$, HM corresponds to the quadratic mean, whereas LM corresponds to the contraharmonic mean. Both families are bounded by the minimum and maximum observations and are non-decreasing functions of $\alpha$ \cite{ziou2024using}.

The parameter $\alpha$ determines the data selection underlying the resulting central tendency. In particular, LM can be expressed as a weighted arithmetic mean:
\begin{equation}
\label{eq:lehmer_weighted}
\hat{Y}_{t+h}=\mathrm{LM}(\alpha,w)
=
\sum_{i=0}^{w-1}
v_i^{\mathrm{LM}}(\alpha)Y_{t-i},
\end{equation}
where
\begin{equation}
\label{eq:lehmer_weights}
v_i^{\mathrm{LM}}(\alpha)
=
\frac{Y_{t-i}^{\alpha-1}}
{\sum_{j=0}^{w-1}Y_{t-j}^{\alpha-1}}.
\end{equation}

Similarly, HM can be expressed as
\begin{equation}
\label{eq:gm_weighted}
\hat{Y}_{t+h}=\mathrm{GM}(\alpha,w)
=
\left(
\sum_{i=0}^{w-1}
v_i^{\mathrm{HM}}(\alpha)Y_{t-i}
\right)^{1/\alpha},
\end{equation}
where
\begin{equation}
\label{eq:gm_weights}
v_i^{\mathrm{HM}}(\alpha)
=
\frac{Y_{t-i}^{\alpha-1}}{w}.
\end{equation}

The two weights illustrate the data selection mechanisms embedded in both families. As $\alpha$ increases, larger observations contribute more strongly to the resulting central tendency, whereas smaller values of $\alpha$ increase the selection of lower observations. Unlike LM, whose weights sum to one, the weights associated with GM do not generally satisfy this property \cite{ziou2024using}.

The mean-based forecasting methods are characterized by two parameters, $\theta_m=(\alpha,w)$, where $m\in\{\mathrm{GM},\mathrm{LM}\}$. To investigate their influence on forecasting accuracy, multiple values of the selection order $\alpha$ and historical window length $w$ are evaluated. For each forecasting horizon $h$ and method $m$, the parameter pair $\theta_{m,h}^{*}=(\alpha_{m,h}^{*},w_{m,h}^{*})$ that minimizes the target forecasting error is selected. Although forecasts generated from the historical observations $Y_{1:t}$ are independent of the forecasting horizon $h$ for a fixed parameter pair $(\alpha,w)$.

\section{Tests and Results}
\label{sec4}
In this section, we evaluate the forecasting performance of the presented methods on the non-holiday dataset presented in Section~\ref{sec2}. As previously described, the dataset is chronologically divided into training, and test sets. In the following, we first describe the experimental configuration and the evaluation metrics. We then investigate the forecasting performance of HL, LM, and TimeGPT considering both one-step-ahead and multi-step-ahead forecasting scenarios.

\subsection{Evaluation Metrics}
\label{sec:metrics}
To assess models performance, we use several common metrics, including mean absolute error (MAE), mean absolute percentage error (MAPE), root mean squared error (RMSE), and residual standard deviation (RSD). These metrics are standard in call forecasting studies and allow for comparisons with existing research \cite{fass2025state,fass2025maroc}. Their interpretation draws inspiration from the work in \cite{ziou2023pythagorean,ziou2024centrality,ziou2024using}. 

The $\text{MAE} = \frac{1}{n} \sum_{i=1}^{n} \left|Y_{i} - \hat{Y}_{i}\right|$ is the AM of the absolute differences between the estimated call volume $\hat{Y}_i$ and the ground truth $Y_i$. By definition of AM, it is sensitive to outliers. For experiments conducted on the same dataset, smaller MAE values indicate more accurate forecasts.

The $\text{RMSE} = \sqrt{\frac{1}{n} \sum_{i=1}^{n} \left(Y_{i} - \hat{Y}_{i}\right)^2}$ is the square root of the AM of the squared differences between the estimated call volume and the ground truth. As a mean, the RMSE is always greater than or equal to the MAE, and it is more sensitive than the MAE to outliers. Like MAE, when using the same data, smaller RMSE values indicate a more accurate forecast.

The $\text{MAPE} = \frac{1}{n} \sum_{i=1}^{n} \left| \frac{Y_{i} - \hat{Y}_{i}}{Y_{i}} \right| \times 100$ measures the relative error, expressed as a percentage of the ground truth.
It is scale-free, making it useful for comparing call volume forecasts regardless of the scale. However, it is undefined when the call volume $Y_{i}$ equals zero. Lower MAPE percentages indicating higher accuracy.

The $\text{RSD} = \sqrt{\frac{1}{n} \sum_{i=1}^{n}{\left( (Y_{i} - \hat{Y}_{i})- ME\right)^2}}$, with $ME=\frac{1}{n}\sum_{i=1}^{n}{\left(Y_{i} - \hat{Y}_{i}\right)}$ measures the dispersion of errors around their mean error ($ME$). Lower RSD values indicate that the errors are more closely grouped, reflecting lower dispersion; for example, the largest errors remain close to the $ME$.

We use also the minimum and maximum absolute errors, denoted by $\mathrm{Min~AE}= \min_{i=1,\ldots,n}|Y_i-\hat{Y}_i|,$ and $\mathrm{Max~AE}=\max_{i=1,\ldots,n}|Y_i-\hat{Y}_i|$, respectively, characterize the extreme values of the absolute forecast errors. Lower extreme values indicate higher accuracy.

Finally, we report the proportions of forecasts  P$_{>\tau}=\frac{100}{n} \sum_{i=1}^{n} \mathbb{I} \left(100\left|\frac{Y_i-\hat{Y}_i}{Y_i}\right|>\tau\right),$ where $\tau\in\{10,20\}$, and $\mathbb{I}(\cdot)$ denotes the indicator function. For the same $\tau$, lower P$_{>\tau}$ values indicate that a smaller proportion of forecasts exceed the corresponding percentage-error threshold.

\subsection{Experimental Setup}
\label{sec:experimental_setup}

For TimeGPT, the training data were divided into a training set covering the period from 2012 to 2015 and a validation set corresponding to 2016. 
Hyperparameter tuning was performed using a grid-search procedure to identify the most suitable fine-tuning configuration for the emergency call forecasting task. Three hyperparameters were considered: the number of fine-tuning steps, the fine-tuning loss function, and the fine-tuning depth. The number of fine-tuning steps, selected from $\{0, 2, 5, 10, 25, 100, 500\}$, controls the number of parameter update iterations during adaptation, with 0 corresponding to the pretrained model without fine-tuning (zero-shot). The fine-tuning loss function, selected from $\{\text{default}, \text{MAE}, \text{MAPE}\}$, determines the objective minimized during parameter updates. The default option uses the loss function provided by TimeGPT, while MAE and MAPE emphasize absolute and relative forecasting errors, respectively. Finally, the fine-tuning depth, selected from $\{1, 2, 3, 4\}$, controls the extent of model adaptation by specifying the number of model layers involved in fine-tuning. Each hyperparameter combination was evaluated on the validation set using MAE as the selection criterion, although other evaluation metrics could also be employed. The configuration yielding the lowest MAE was retained and subsequently used to fine-tune TimeGPT on the combined training and validation sets before its final evaluation on the test set.

For the HM and LM models, a grid search was conducted over the parameter $\alpha \in [-10, 10]$, with a step size of $0.05$, and historical window sizes $w \in \{7, 14, 30, \ldots,|Y_{1:t}|\}$. All configurations were evaluated on the test set.

Unless otherwise stated, all forecasting performance metrics reported in the following were computed on the test set.

\subsection{Discussion}
\label{subsec:results}
Table~\ref{tab:forecast_comparison} describes the forecasting performance of univariate TimeGPT, HM, and LM on the test dataset for three forecasting horizons, $h=1$, $h=3$, and $h=7$. For one-step-ahead ($h=1$) and three-step-ahead ($h=3$) forecasting, both generalized means outperform TimeGPT across most evaluation metrics, except Max AE. Specifically, HM and LM achieve MAE values of approximately 25.065 and 25.790, compared with 25.380 and 26.446 for TimeGPT, corresponding to reductions of 1.24\% and 2.48\%, respectively. Similarly, the generalized means achieve lower MAPE values (8.293\% and 8.489\%) than TimeGPT (8.462\% and 8.838\%). Improvements are also observed in RMSE, RSD, and the proportion of predictions with relative errors exceeding 10\% and 20\%. However, TimeGPT exhibits lower maximum absolute errors at both horizons, with values of 90.991 and 95.255, compared with approximately 95.089 and 100.429 for the generalized means. For seven-step-ahead forecasting ($h=7$), the ranking is reversed, with TimeGPT achieving the lowest error in terms of all metrics except Max AE. The MAE of TimeGPT is approximately 1.98\% lower than that of the generalized means.
\begin{table*}[t]
\centering
\caption{Forecasting performance comparison on the test dataset. Values in parentheses indicate 95\% bootstrap confidence intervals. Best scores for each horizon are highlighted in bold.}
\label{tab:forecast_comparison}
\footnotesize

\resizebox{\textwidth}{!}{%
\begin{tabular}{lrrrrrrrr}
\hline
\textbf{Model} &
\textbf{MAE (95\% CI)} &
\textbf{MAPE (\%) (95\% CI)} &
\textbf{RMSE} &
\textbf{RSD} &
\textbf{Min AE} &
\textbf{Max AE} &
\textbf{P$_{>10\%}$} &
\textbf{P$_{>20\%}$} \\
\hline

\multicolumn{9}{l}{\textbf{($h=1$)}} \\
\hline

Univariate TimeGPT &
25.380 (23.514, 27.309) &
8.462 (7.869, 9.067) &
31.916 & 31.868 &
0.146 & \textbf{90.991} & 32.479 & 7.407 \\

HM &
25.067 (23.082, 26.989) &
8.294 (7.628, 8.971) &
31.495 & 31.485 &
0.015 & 95.089 & \textbf{31.250} & \textbf{6.534} \\

LM &
\textbf{25.065} (23.081, 26.989) &
\textbf{8.293} (7.627, 8.971) &
\textbf{31.493} & \textbf{31.482} &
\textbf{0.000} & 95.079 &
\textbf{31.250} & 6.818 \\

\hline

\multicolumn{9}{l}{\textbf{($h=3$)}} \\
\hline

Univariate TimeGPT &
26.446 (24.391, 28.577) &
8.838 (8.119, 9.587) &
33.203 & 32.990 &
0.254 & \textbf{95.255} & 35.429 & 8.286 \\

HM &
\textbf{25.790} (23.765, 27.785) &
\textbf{8.489} (7.824, 9.191) &
\textbf{32.468} & \textbf{32.460} &
\textbf{0.214} & 100.429 &
\textbf{33.523} & \textbf{7.670} \\

LM &
\textbf{25.790} (23.765, 27.785) &
\textbf{8.489} (7.824, 9.191) &
\textbf{32.468} & \textbf{32.460} &
\textbf{0.214} & 100.429 &
\textbf{33.523} & \textbf{7.670} \\

\hline

\multicolumn{9}{l}{\textbf{($h=7$)}} \\
\hline

Univariate TimeGPT &
\textbf{25.730} (23.619, 27.863) &
\textbf{8.562} (7.912, 9.243) &
\textbf{32.504} & \textbf{32.481} &
0.308 & 101.898 &
\textbf{31.909} & \textbf{7.407} \\

HM &
\textbf{}26.251 (24.099, 28.371) &
8.634 (7.952, 9.351) &
33.614 & 33.612 &
\textbf{0.029} & \textbf{98.437} &
34.659 & 7.670 \\

LM &
26.251 (24.100, 28.371) &
8.635 (7.953, 9.351) &
33.613 & 33.611 &
0.052 & 98.453 &
34.659 & 7.670 \\

\hline
\end{tabular}%
}
\end{table*}

\begin{figure}[t]
    \centering
    \includegraphics[width=0.8\linewidth]{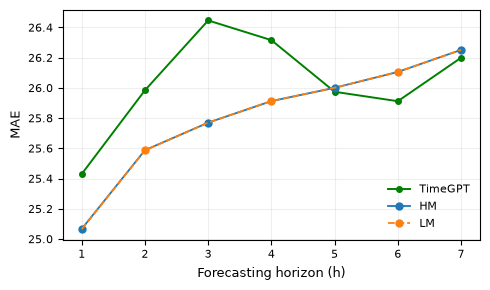}
    \caption{Forecasting performance as a function of the
    prediction horizon for TimeGPT, HM, and LM.}
    \label{fig:horizon_comparison}
\end{figure}

To investigate the influence of the forecasting horizon, Fig.~\ref{fig:horizon_comparison} illustrates the evolution of the MAE across forecasting horizons from $h=1$ to $h=7$. Although MAE is used for this analysis, a similar comparison can be conducted using other evaluation metrics. The scores show that HM and LM exhibit nearly identical forecasting performance across all horizons, with their MAE increasing progressively from approximately 25.07 at $h=1$ to 26.25 at $h=7$. This corresponds to an increase of approximately 4.73\%, indicating a gradual deterioration in forecasting accuracy as the prediction horizon increases. In contrast, TimeGPT exhibits a smaller increase in MAE, from 25.38 at $h=1$ to 25.73 at $h=7$, corresponding to approximately 1.38\%. For horizons $h=1$ to $h=4$, both generalized means outperform TimeGPT, while at $h=5$, all three models achieve comparable MAE. However, for longer horizons ($h=6$ and $h=7$), TimeGPT achieves a lower MAE than both generalized means. 

To further assess whether the observed differences in forecasting performance are statistically significant, pairwise Diebold--Mariano (DM) tests were conducted \cite{Diebold01071995}. The DM test compares the predictive accuracy of two forecasting methods by examining the difference between their forecasting losses. In this study, absolute errors were used as the loss function, and the null hypothesis assumes that both methods have equal expected forecasting losses. The tests were applied to TimeGPT, HM, and LM for forecasting horizons $h=1$ and $h=7$. Since forecasting errors may be correlated over time, this temporal dependence was accounted for using a heteroskedasticity and autocorrelation consistent (HAC) variance estimator with Bartlett weights \cite{newey1986simple}. Table~\ref{tab:dm_comparison} reports the corresponding p-values, with a significance threshold of 5\%. The scores show that all p-values exceed the significance threshold of 0.05, indicating that the DM tests do not provide sufficient evidence to reject the null hypothesis of equal predictive accuracy at either forecasting horizon. Consequently, the observed numerical differences should not be interpreted as statistically significant improvements. Overall, the results suggest that despite being pretrained on approximately 100 billion observations and using hundreds of millions of parameters, TimeGPT does not outperform training-free mean-based method. These findings highlight the strong potential of frugal methods as alternatives to energyvore methods.
\begin{table}[t]
\centering
\caption{Pairwise Diebold--Mariano test p-values based on absolute forecast errors.}
\label{tab:dm_comparison}
\footnotesize
\begin{tabular}{p{1.5 cm}p{3.3cm}p{3.3cm}p{3.3cm}}
\hline
\textbf{Horizon} &
\textbf{TimeGPT vs HM} &
\textbf{TimeGPT vs LM} &
\textbf{HM vs LM} \\
\hline
$h=1$ & 0.4793 & 0.4765 & 0.1209 \\
$h=7$ & 0.5597 & 0.5594 & 0.8556 \\
\hline
\end{tabular}
\end{table}

\begin{figure*}[t]
    \centering
    \begin{subfigure}[t]{0.49\textwidth}
        \centering
        \includegraphics[width=\linewidth]{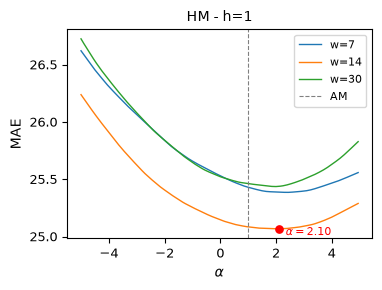}
        \caption{HM ($h=1$)}
        \label{fig:hm_h1}
    \end{subfigure}
    \begin{subfigure}[t]{0.49\textwidth}
        \centering
        \includegraphics[width=\linewidth]{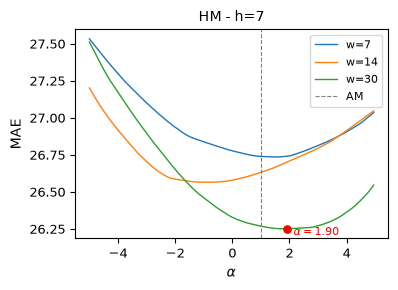}
        \caption{HM ($h=7$)}
        \label{fig:hm_h7}
    \end{subfigure}
    \\
    \begin{subfigure}[t]{0.49\textwidth}
        \centering
        \includegraphics[width=\linewidth]{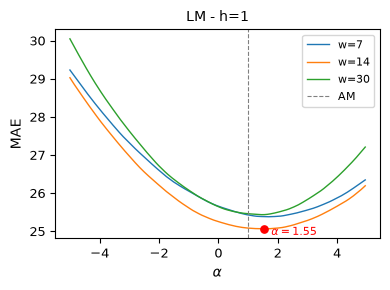}
        \caption{LM ($h=1$)}
        \label{fig:lm_h1}
    \end{subfigure}
    \begin{subfigure}[t]{0.49\textwidth}
        \centering
        \includegraphics[width=\linewidth]{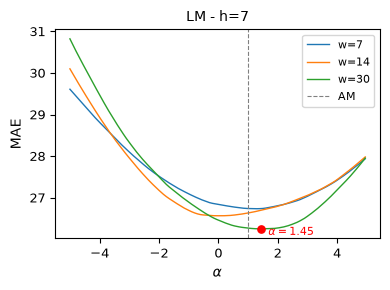}
        \caption{LM ($h=7$)}
        \label{fig:lm_h7}
    \end{subfigure}
    \caption{Effect of the parameter $\alpha$ and historical window size $w$ on forecasting performance for HM and LM at horizons $h=1$ and $h=7$.}
    \label{fig:generalized_means}
\end{figure*}

To investigate the sensitivity of HM and LM to their hyperparameters, Fig.~\ref{fig:generalized_means} illustrates the variation in MAE as a function of $\alpha$ for different historical window sizes $w$ at forecasting horizons $h=1$ and $h=7$, while Table~\ref{tab:optimal_parameters} reports the optimal configurations across all horizons from $h=1$ to $h=7$. Although MAE is used for this analysis, other evaluation metrics lead to similar conclusions. Interestingly, both generalized means select the same optimal window size for each forecasting horizon. For shorter horizons ($h=1,2,3$), a window of 14 observations is preferred, whereas a longer window of 30 observations is selected for horizons $h=4,\ldots,7$. These suggest that the number of observations used to generate forecasts plays an important role in their predictive performance.
In contrast, the optimal values of $\alpha$ differ between the two models. For HM, $\alpha$ ranges from 0.05 to 2.10, whereas for LM it ranges from 0.55 to 1.55, suggesting that the AM ($\alpha=1$) is not necessarily the optimal choice for this forecasting task. In particular, for $h=1$, the optimal configurations are $(\alpha,w)=(2.10,14)$ for HM and $(1.55,14)$ for LM, whereas for $h=7$, they shift to $(1.90,30)$ and $(1.45,30)$, respectively. These differences can be attributed to their distinct mathematical formulations, in which $\alpha$ controls the data selection differently. Nevertheless, despite selecting different values of $\alpha$, HM and LM achieve nearly identical errors across all forecasting horizons. 
\begin{table}[t]
\centering
\caption{Optimal values of $\alpha$ and window size $w$ for HM and LM across forecasting horizons.}
\label{tab:optimal_parameters}
\footnotesize
\begin{tabular}{p{1.5cm} p{1cm}p{1cm}p{1cm}p{1cm}}
\hline
\textbf{Horizon} &
\multicolumn{2}{l}{\textbf{HM}} &
\multicolumn{2}{l}{\textbf{LM}} \\
\cline{2-5}
\textbf{$h$   } &
\textbf{$\alpha$   } &
\textbf{$w$} &
\textbf{$\alpha$} &
\textbf{$w$} \\
\hline
1 & 2.10 & 14 & 1.55 & 14 \\
2 & 1.60 & 14 & 1.30 & 14 \\
3 & 0.05 & 14 & 0.55 & 14 \\
4 & 1.45 & 30 & 1.20 & 30 \\
5 & 1.35 & 30 & 1.20 & 30 \\
6 & 1.75 & 30 & 1.30 & 30 \\
7 & 1.90 & 30 & 1.45 & 30 \\
\hline
\end{tabular}
\end{table}

Overall, these findings highlight the importance of jointly selecting the weighting parameter $\alpha$ and the historical window size $w$ according to the forecasting horizon. They also demonstrate that HM and LM exhibit remarkably similar predictive behavior despite their different formulations, supporting the potential of generalized means as simple forecasting approaches for daily emergency call volume prediction.

\section{Conclusion}
\label{sec5}
This study investigated the potential of generalized means as simple alternatives to the TimeGPT foundation model for daily emergency call volume forecasting. Two families of generalized means, the HM and the LM, were evaluated across forecasting horizons ranging from one to seven days. Their predictive performance was compared with that of fine-tuned TimeGPT using MAE, MAPE, RMSE, RSD, Max AE, Min AE, $\text{P}_{\tau}$ metrics and Diebold--Mariano tests. The experimental results showed that for short-term forecasting, both generalized means outperformed TimeGPT in terms of MAE, achieving reductions of approximately 1.24\% and 2.48\% at horizons of one and three days, respectively. However, TimeGPT achieved better performance at longer horizons, with an MAE approximately 1.98\% lower than that of the generalized means at the seven-day horizon. Despite these numerical differences, the Diebold--Mariano tests revealed no statistically significant differences in predictive accuracy between the three methods at the evaluated horizons of one and seven days. Furthermore, the results highlighted the influence of the aggregation parameter $\alpha$ and historical window size $w$ on the predictive performance of generalized means. In particular, shorter historical windows were preferred for short-term forecasts, whereas longer windows were selected for extended forecasting horizons.

Overall, these findings suggest that generalized means can achieve forecasting performance comparable or better than TimeGPT. Future research will investigate the potential benefits of incorporating explanatory variables into generalized mean-based forecasting approaches.

\bibliography{sn-bibliography}

@misc{ccsll,
    title = {Billan Annuel 2021-2022, Centre de communication santé Laurentides + Lanaudière},
    author={CCSLL},
    year={2022},
    howpublished ={URL : \url{https://ccsll.org/wp-content/uploads/Bilan_Annuel_2021-2022_CCSLL.pdf}}, 
    
}

@conference{fass2026TimeGPT,
    author = {Feriel Fass and Djemel Ziou} ,
    title = {Daily Emergency Call Volume Forecasting using Univariate and Multivariate TimeGPT}, 
    year = {2026},
    booktitle={The 2nd International Conference on Health Medical Systems and Services and Services (HMSS 2026)},
}

@article{ibrahim2016modeling,
  title={Modeling and forecasting call center arrivals: A literature survey and a case study},
  author={Ibrahim, Rouba and Ye, Han and L’Ecuyer, Pierre and Shen, Haipeng},
  journal={International Journal of Forecasting},
  volume={32},
  number={3},
  pages={865--874},
  year={2016},
  publisher={Elsevier}
}

@article{fildes2002telecommunications,
  title={Telecommunications demand forecasting—a review},
  author={Fildes, Robert and Kumar, V},
  journal={International journal of forecasting},
  volume={18},
  number={4},
  pages={489--522},
  year={2002},
  publisher={Elsevier}
}

@article{fass2025state,
author = {Fass, Feriel and Mecheri, Hadia and Ziou, Djemel and L\'{e}vesque, Jessica},
title = {State-of-the-Art Review and Comparative Experimentation of Emergency Call Prediction Models},
year = {2026},
publisher = {Association for Computing Machinery},
address = {New York, NY, USA},
volume = {58},
number = {8},
issn = {0360-0300},
url = {https://doi.org/10.1145/3785661},
doi = {10.1145/3785661},
journal = {ACM Comput. Surv.},
month = feb,
articleno = {199},
numpages = {19},
}

@article{fass2025survey,
title = {Computational models for emergency prehospital care},
journal = {Computer Science Review},
volume = {63},
pages = {101077},
year = {2027},
issn = {1574-0137},
doi = {https://doi.org/10.1016/j.cosrev.2026.101077},
url = {https://www.sciencedirect.com/science/article/pii/S1574013726001851},
author = {Feriel Fass and Djemel Ziou and Jessica Lévesque},
}

@misc{ziou2023pythagorean,
      title={Pythagorean Centrality for Data Selection}, 
      author={Djemel Ziou},
      note={arXiv preprint},
      year={2023},
      eprint={2301.10010},
      archivePrefix={arXiv},
      primaryClass={math.HO},
      howpublished={https://arxiv.org/abs/2301.10010}, 
}

@misc{ziou2024centrality,
      title={Centrality Estimators for Probability Density Functions}, 
      author={Djemel Ziou},
      note={arXiv preprint},
      year={2024},
      eprint={2404.05816},
      archivePrefix={arXiv},
      primaryClass={math.ST},
      howpublished={https://arxiv.org/abs/2404.05816}, 
}

@misc{benamirouche2026,
      title={A Transferable Autologistic Model for Predicting Rare Failures in Heterogeneous Equipment}, 
      author={Islam Benamirouche and Djemel Ziou and Feriel Fass},
      note={arXiv preprint},
      year={2026},
      eprint={2608.06695},
      archivePrefix={arXiv},
      primaryClass={cs.LG},
      howpublished={https://arxiv.org/abs/2608.06695}, 
}

@article{newey1986simple,
 ISSN = {00129682, 14680262},
 URL = {http://www.jstor.org/stable/1913610},
 author = {Whitney K. Newey and Kenneth D. West},
 journal = {Econometrica},
 number = {3},
 pages = {703--708},
 publisher = {[Wiley, Econometric Society]},
 title = {A Simple, Positive Semi-Definite, Heteroskedasticity and Autocorrelation Consistent Covariance Matrix},
 urldate = {2026-10-29},
 volume = {55},
 year = {1987}
}

@article{ziou2024using,
  title={Using Maximum Weighted Likelihood to Derive Lehmer and H{\"o}lder Families of Means},
  author={Ziou, Djemel},
  journal={IEEE Signal Processing Letters},
  volume={31},
  number={},
  pages={1625--1629},
  year={2024},
  publisher={IEEE}
}

@InProceedings{fass2025maroc,
author="Fass, Feriel
and Mecheri, Hadia
and Ziou, Djemel
and  L{\'e}vesque, Jessica",
title="Predicting Call Volume Using Coevolutionary Phenomena and Stochastic Models",
booktitle="Artificial Intelligence and Green Computing",
year="2025",
publisher="Springer Nature Switzerland",
address="Cham",
pages="360--371",
isbn="978-3-032-02312-4"
}

@article{Diebold01071995,
author = {Francis X. Diebold and Roberto S. Mariano},
title = {Comparing Predictive Accuracy},
journal = {Journal of Business \& Economic Statistics},
volume = {13},
number = {3},
pages = {253--263},
year = {1995},
publisher = {Taylor \& Francis},
doi = {10.1080/07350015.1995.10524599},
URL = { https://www.tandfonline.com/doi/abs/10.1080/07350015.1995.10524599},
eprint = { https://www.tandfonline.com/doi/pdf/10.1080/07350015.1995.10524599}
}

@misc{garza2024timegpt1,
      title={TimeGPT-1}, 
      author={Azul Garza and Cristian Challu and Max Mergenthaler-Canseco},
      note={arXiv preprint},
      year={2024},
      eprint={2310.03589},
      archivePrefix={arXiv},
      primaryClass={cs.LG},
      howpublished={https://arxiv.org/abs/2310.03589}, 
}
\bibliographystyle{ieeetr}

\end{document}